# Auto Prompt SQL: A Resource-Efficient Architecture for Text-To-SQL Translation in Constrained Environments

*Zetong Tang, Qian Ma, Di Wu**

*College of Computer and Information Science Southwest University, Chongqing, China*
** wudi1986@swu.edu.cn*

**Keywords**: Language Model, Text-to-SQL, Prompt Engineering, Fine-tuning, In-Context Learning

## Abstract

Using the best Text-to-SQL methods in resource-constrained environments is challenging due to their reliance on resource-intensive open-source models. This paper introduces Auto Prompt SQL(AP-SQL), a novel architecture designed to bridge the gap between resource-efficient small open-source models and the powerful capabilities of large closed-source models for Text-to-SQL translation. Our method decomposes the task into schema filtering, retrieval-augmented text-to-SQL generation based on in-context examples, and prompt-driven schema linking and SQL generation. To improve schema selection accuracy, we fine-tune large language models. Crucially, we also explore the impact of prompt engineering throughout the process, leveraging Chain-of-Thought(CoT) and Graph-of-Thought(GoT) templates to significantly enhance the model's reasoning for accurate SQL generation. Comprehensive evaluations on the Spider benchmarks demonstrate the effectiveness of AP-SQL.

## 1 Introduction

The task of Text-to-SQL involves translating questions into executable and valid Structured Query Language (SQL) queries. It demonstrates the process of formulating natural language questions based on a database schema and subsequently transforming them into SQL queries[4,5].

*1.1 Limitations and Shortcomings of the Latest Technology*

Traditional approaches predominantly employed supervised fine-tuning. However, the recent emergence of a series of large language models, including GPT, Claude, Qwen, and Kimi, has initiated a paradigm shift, with prompt engineering beginning to gain prominence in this domain. (1) Large closed-source models exhibit strong reasoning abilities stemming from their extensive parameter training. These parameters can potentially influence the reasoning process of large models, leading to the generation of hallucinated information[6]. (2) The knowledge of these models is limited to the time point of their training data, resulting in a lack of accurate understanding of unseen knowledge and information[9,10,21]. (3) The opacity of the reasoning process makes it difficult to explain how the model arrives at specific conclusions and prevents the verification of the model's reasoning path[7]. This paper introduces Auto-prompt, a plug-and-play open-source language model designed for large models. The pattern selection module within this framework is constructed based on fine-tuning the Qwen model and employs a large language model for pattern linking classification. Finally, text-to-SQL generation is achieved by integrating with a generative large model through our designed prompt-prompt template.

*1.2 Research Status*

In our investigation, we covered supervised fine-tuning and prompt engineering-based methods for text-to-SQL. Furthermore, we also investigated various existing schema linking techniques and prompt design techniques that can be used to enhance text-to-SQL methods[25-70].

*1.2.1 Supervised Fine-tuning for Text-to-SQL:* Prior to the advent of large language model technology, the dominant approach for text-to-SQL was fine-tuning 'encoder-decoder' neural network models[11,12,14,15]. Other endeavors have concentrated on injecting SQL grammar into the decoder, which constrains the decoder's output space, ensuring the generation of syntactically correct SQL queries[8]. With the advancement of language models, there has been a growing trend towards formulating text-to-SQL as a sequence-to-sequence (seq2seq) task[23,24].

*1.2.2 Prompt Engineering-based Text-to-SQL:* With the emergence of large language models, such as the ChatGPT series, Claude series, Qwen series, and Llama series, a revolutionary transformation has been brought to the field of Natural Language Processing (NLP). These models have achieved remarkable progress on various complex tasks requiring reasoning, without the need for fine-tuning any parameters. Leveraging text-to-SQL demonstrations as a few-shot prompt, they have successfully achieved state-of-the-art (SOTA) performance on the Spider benchmark[13,17,19].

*1.2.3 Schema Link:* Schema linking plays a crucial role in the text-to-SQL process, a task that aims to identify database schemas (tables and columns) and database values referenced in a natural language question. There are two main strategies for schema linking: string matching-based and neural network-based. String matching-based methods appear simple and effective, but they have limitations when dealing with synonyms and polysemous words[22].

*1.2.4 Retrieval-Augmented Generation(RAG):* It essentially employs engineered methods to address the limitations of large language models. Its core mechanism involves utilizing an external knowledge database connected to the large language model to store new data, domain-specific data, etc., that were not present in the training data[16,18,20].

## 2 Methodology

The goal of Text-to-SQL is to generate an SQL query $S$ based on a natural language question $Q$ and database schema information $D$, the process can be formulated as follows:

$$S = P(Q,D) \tag{1}$$

the model $P$ required to match and understand the natural language question $Q$ with the database schema information $D$, and then generate the SQL statement.

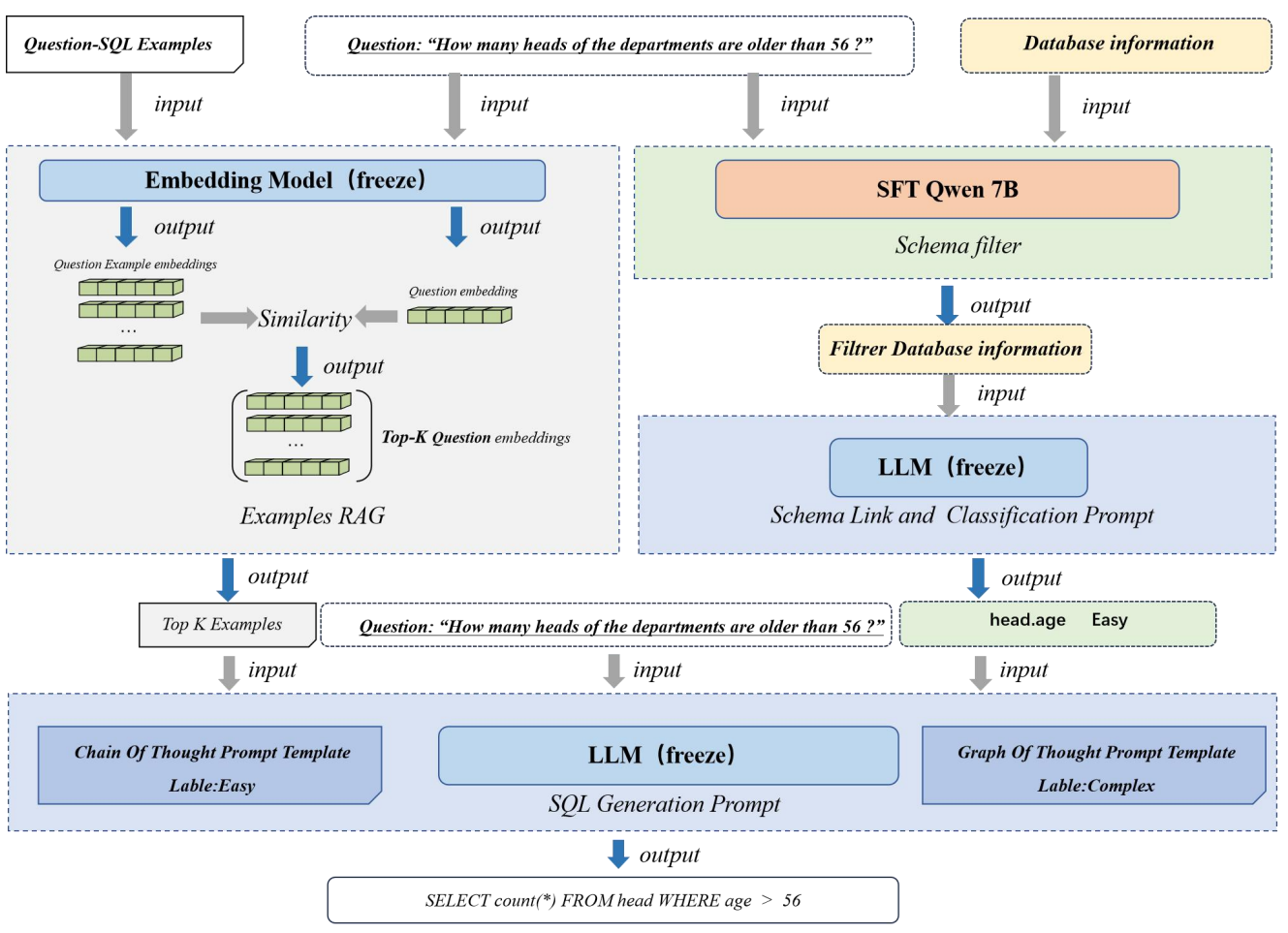

Fig. 1. Illustration of the comprehensive framework

The framework of Auto-Prompt is shown in Fig.1, first, we construct a schema filter using a fine-tuned large language model to efficiently collect table and column information relevant to the question. Subsequently, combining a schema linking prompting strategy, we guide the model to complete the schema linking task and design a difficulty grading mechanism for different questions. To further enhance the reasoning ability of the large model, we introduce an examples RAG module, which selects the Top-K most relevant NL-SQL examples from the training set for reasoning enhancement. In the final SQL generation stage, we design chain-of-thought prompting templates and thought graph prompting templates to guide the large model to perform structured and hierarchical reasoning generation, thereby improving the accuracy of SQL generation.

Using prompt engineering methods, the SQL generation process can be formulated as follows:

$$S = \mathrm{LM}(I,D,Q,\mathcal{E}) \tag{2}$$

$I$ represents the instruction, $D$ represents the database schema information, and $Q$ represents the question. $\mathcal{E} = [(D_1,Q_1,P_1),\ldots, D_n,Q_n,P_n)]$ represents sample list. where $P_i$ is the correct SQL answer prompt for the i-th sample. Therefore, the SQL generation performance of the large model is primarily influenced by the accuracy of the schema information, the selection of samples, and the design of the prompt style.

### *2.1 Examples Retrieval-Augmented Generation*

To improve the cross-domain adaptability of the Text-to-SQL framework and enhance large language model performance in few-shot settings without complex fine-tuning, we developed a RAG module. This module dynamically supplies external knowledge relevant to the input question for schema linking and SQL generation. Specifically, it retrieves the Top-K most relevant Text-SQL pairs from a pre-built, high-quality example library based on the user's natural language question. These examples, combined with the question, are fed into the model as contextual prompts, enabling better semantic understanding and more accurate SQL query generation.

### *2.2 Schema Filtering*

In practical applications, databases with numerous tables and columns result in lengthy prompts for large language models, increasing computational load and risking oversight of critical tables and fields.

To address this, we propose a question-based schema filtering method to compress prompt length while preserving key information. Specifically, given a natural language question, our method selects the top 3 most relevant tables from the database and retains the top 3 most relevant columns per table, significantly reducing prompt length while maintaining schema information closely tied to the question. The filtering model, based on a supervised fine-tuned Qwen-3B, excels in schema understanding and question association, enabling precise filtering. It was fine-tuned on a dataset of approximately 80,000 annotated question-schema pairs over 10 epochs, using a learning rate of 5e-5 and a batch size of 16. The training was conducted on 2 4090D GPUs, ensuring sufficient computational resources for stable convergence and performance.

### *2.3 Schema Linking*

In our constructed schema linking prompt design, we introduced a few-shot learning mechanism to guide the model's question reasoning. The entire process sequentially goes through two stages: table selection and column selection, progressing step-by-step based on the filtered schema information, and finally selecting the table and column information relevant to the current question.

Furthermore, in the table selection process, we also explored leveraging the large model itself to evaluate its progress in addressing the problem. Specifically, we independently scored each table based on the question to determine its relevance. The process can be formulated as follows:

$$\forall d \in D_{tables}, V(p,q) \sim (d \mapsto v(d)) \tag{3}$$

An individual table $d$ from the set of tables in the database schema $D_{tables}$, $V(p,q)$ is a function that evaluates the relevance of a table $d$ given the prompt $p$ and question $q$. $v(d)$ is scalar value(1-10), select the information of tables with a score greater than a certain threshold (e.g., 6). After completing this step, we then use voting to compare the information of different columns in d.

*2.4 SQL Generation*

To address SQL generation tasks, we employ tailored prompting strategies based on query complexity. For simple single-table queries involving basic operations such as filtering, sorting, and aggregation, we adopt chain-of-thought (CoT) prompting, which guides the model through a linear, step-by-step reasoning process.

CoT prompting promotes structured reasoning by presenting exemplar problem-solution pairs, each consisting of a natural language question, a detailed reasoning process, and the final SQL query. This method helps the model generalize to new queries by leveraging a small set of carefully selected examples. For more complex multi-table queries—such as those involving nested subqueries, conditional joins, grouping, and advanced SQL constructs—we use thought graph prompting. This method models problem-solving as a graph of interconnected reasoning nodes. Each exemplar consists of a question, a structured graph detailing intermediate reasoning steps, and the final solution, allowing the model to systematically decompose and solve intricate queries.

# 3 Results

*3.1 Datasets*

We primarily conducted the main Text-to-SQL experiments on the Spider dataset, which contains 8,659 training samples and 1,034 development samples. Spider spans 200 databases across 138 different domains.

For models evaluated on Spider, we primarily used Execution Accuracy and Test Suite Accuracy. EX evaluates whether the predicted SQL and the ground truth SQL query produce the same execution results on the database. Due to the possibility of false positives in Execution Accuracy, where an incorrect SQL query happens to produce the same output as the correct one, we use TS to evaluate whether the generated SQL query consistently passes the EX evaluation across multiple database instances. The experiments was conducted on 2 4090D GPUs, the schema linking threshold was set to 6, and the RAG module retrieved $K$ =3 Text-SQL pairs.

*3.2 Comparative Experiment*

Table 1 clearly shows that AP-SQL, on the Spider dataset, consistently outperforms other models across various LLMs, including Qwen-7B, Llama-8B, GPT-4o-mini, and GPT-4o. For each model, AP-SQL achieves the highest EX% and TS% scores, with notable improvements over E-SQL[3], ACT-SQL[1], and C3-SQL[2].

Table 1. Evaluation of AP-SQL on Spider.

| Methods | LLM | EX% | TS% |
|---|---|---|---|
| E-SQL | Qwen-7B | 67.8 | 60.4 |
| DIN-SQL | | 63.7 | 54.6 |
| ACT-SQL | | 51.1 | 43.3 |
| C3-SQL | | 52.6 | 47.5 |
| **AP -SQL (ours)** | | **68.3** | **60.8** |
| E-SQL | Llama-8B | 70.2 | 63.3 |
| DIN-SQL | | 69.5 | 61.4 |
| ACT-SQL | | 63.7 | 55.5 |
| C3-SQL | | 60.9 | 50.2 |
| **AP -SQL (ours)** | | **72.4** | **64.1** |
| E-SQL | GPT-4o-mini | 82.6 | 72.4 |
| DIN-SQL | | 74.2 | 69.6 |
| ACT-SQL | | 80.7 | 70.2 |
| C3-SQL | | 78.1 | 71.4 |
| **AP-SQL (ours)** | | **83.2** | **75.8** |
| E-SQL | GPT-4o | 88.6 | 79.4 |
| DIN-SQL | | 87.6 | 81.5 |
| ACT-SQL | | 86.7 | 78.2 |
| C3-SQL | | 86.2 | 79.6 |
| **AP -SQL (ours)** | | **89.7** | **82.6** |

This consistent superiority highlights AP-SQL's effectiveness in enhancing SQL query generation across diverse language models. Compared with DIN-SQL, AP-SQL significantly improves the adaptability of complex database schemas and reduces inference costs by decoupling schema links and generation modules.

# 4 Conclusion

Auto-Prompt, through its modular design, the use of supervised fine-tuning, and prompt engineering, significantly enhances Text-to-SQL performance. Its open-source nature, strong adaptability to new domains, and efficient inference make it suitable for low-resource environments. Future work could further optimize prompt design to support more complex database queries.

# 5 Acknowledgements

This work was supported by the Science and Technology Foundation of State Grid Corporation of China under grant 1400-202357341A-1-1-ZN (Identification of Energy Security Risks and Strategic Path Optimization Technology Research under Global Coal-Oil-Gas-Electricity Coupling in China).

# 6 References

[1]. Dong X, Zhang C, Ge Y, et al. :'C3: Zero-shot text-to-SQL with ChatGPT'. arXiv preprint arXiv:2307.07306, 2023

[2]. Caferoğlu, H. A., Ulusoy, Ö. :'E-sql: Direct schema linking via question enrichment in text-to-sql'. arXiv preprint arXiv:2409.16751, 2024

[3]. Li, H., Zhang, J., Liu, H., et al.:'Codes: Towards building open-source language models for text-to-SQL', Proc. ACM Manage. Data, 2024, 2, (3), pp. 1–28
[4]. Gao, Y., Liu, Y., Li, X., et al.:'Xiyan-SQL: A multi-generator ensemble framework for text-to-SQL'. arXiv preprint arXiv:241 unter1.08599, 2024
[5]. Li, H., Zhang, J., Li, C., et al.:'RESDSQL: Decoupling schema linking and skeleton parsing for text-to-SQL'. Proc. AAAI Conf. Artificial Intelligence, Washington, D.C., USA, February 2023, AAAI Press, 37, (11), pp. 13067–13075
[6]. Wang, B., Shin, R., Liu, X., et al.:'RAT-SQL: Relation-Aware Schema Encoding and Linking for Text-to-SQL Parsers'. Proc. 58th Annu. Meet. Assoc. Comput. Linguistics, Virtual, July 2020, Association for Computational Linguistics, pp. 7567–7578
[7]. Luo, X., Wu, H., Wang, Z., et al.:'A Novel Approach to Large-Scale Dynamically Weighted Directed Network Representation', IEEE Trans. Pattern Anal. Mach. Intell., 2022, 44, (12), pp. 9756–9773
[8]. Wu, D., Luo, X., Shang, M., et al.:'A Data-Characteristic-Aware Latent Factor Model for Web Services QoS Prediction', IEEE Trans. Knowl. Data Eng., 2022, 34, (6), pp. 2525–2538
[9]. Shang, M., Yuan, Y., Luo, X., et al. :'An α-β-divergence-generalized Recommender for Highly-accurate Predictions of Missing User Preferences', IEEE Transactions on Cybernetics, 2022, 52,(8),pp. 8006-8018.
[10]. Wu, D., He, Y., Luo, X., et al. :'A Latent Factor Analysis-based Approach to Online Sparse Streaming Feature Selection'. IEEE Transactions on Systems Man Cybernetics: Systems, 2022, 52,(11),pp. 6744-6758.
[11]. Luo, X., Wang, D., Zhou, M., et al. :'Latent Factor-based Recommenders Relying on Extended Stochastic Gradient Descent Algorithms'. IEEE Transactions on Systems Man Cybernetics: Systems, 2021, 51,(2),pp. 916-926.
[12]. Li,W., He,Q., Luo, X. , et al. :'Assimilating Second-Order Information for Building Non-Negative Latent Factor Analysis-Based Recommenders'. IEEE Transactions on Systems Man Cybernetics: Systems, 2021, 52,(1),pp. 485-497.
[13]. Luo, X., Zhou,M., Li,S. :'Algorithms of Unconstrained Non-negative Latent Factor Analysis for Recommender Systems', IEEE Transactions on Big Data, 2021, 7,(1),pp. 227-240.
[14]. Luo, X., Zhou,M. :'Effects of Extended Stochastic Gradient Descent Algorithms on Improving Latent Factor-based Recommender Systems'. IEEE Robotics and Automation Letters, 2019, 4,(2),pp. 618-624.
[15]. Wu,D., Luo, X., Wang,G., et al. :'A Highly-Accurate Framework for Self-Labeled Semi-Supervised Classification in Industrial Applications'. IEEE Transactions on Industrial Informatics, 2018, 14,(3),pp. 909-920.
[16]. Luo, X., Sun,J., Wang,Z., et al. :'Symmetric and Non-negative Latent Factor Models for Undirected, High Dimensional and Sparse Networks in Industrial Applications'. IEEE Transactions on Industrial Informatics, 2017, 13,(6),pp. 3098-3107.
[17]. Luo, X., Zhou,M., Li,S., et al. :'A Nonnegative Latent Factor Model for Large-Scale Sparse Matrices in Recommender Systems via Alternating Direction Method'. IEEE Transactions on Neural Networks and Learning Systems, 2016, 27,(3),pp. 524-537.
[18]. Luo, X., Zhou,M, Li,S., et al. :'An Efficient Second-order Approach to Factorizing Sparse Matrices in Recommender Systems'. IEEE Transactions on Industrial Informatics. 2015, 11,(4),pp. 946-956.
[19]. Luo, X., Zhou,M., Xia,Y. , et al. :'An Efficient Non-negative Matrix-factorization-based Approach to Collaborative-filtering for Recommender Systems'. IEEE Transactions on Industrial Informatics, 2014, 10,(2),pp. 1273-1284.
[20]. Wu,D., He,Y., Luo, X. :'A Graph-incorporated Latent Factor Analysis Model for High-dimensional and Sparse Data'. IEEE Transactions on Emerging Topics in Computing, 2023, 11,(4),pp. 907-917.
[21]. Wu,D., Zhang,P., He,Y., et al. :'A Double-Space and Double-Norm Ensembled Latent Factor Model for Highly Accurate Web Service QoS Prediction', IEEE Transactions on Services Computing, 2023, 16,(2),pp. 802-814.
[22]. Luo, X., Liu,Z., Shang,M., et al. :'Highly-Accurate Community Detection via Pointwise Mutual Information-Incorporated Symmetric Non-negative Matrix Factorization', IEEE Transactions on Network Science and Engineering, 2021, 8,(1), pp.463-476.
[23]. Li,Z., Li,S., Luo, X. :'An Overview of Calibration Technology of Industrial Robots', IEEE/CAA Journal of Automatica Sinica, 2021, 8,(1),pp.23-36.
[24]. Luo, X., Zhou,M., Xia,Y. , et al. :'An Efficient Non-negative Matrix-factorization-based Approach to Collaborative-filtering for Recommender Systems'. IEEE Transactions on Industrial Informatics, 2014, 10,(2),pp. 1273-1284.
[25]. Wu,D., Luo, X., He,Y., et al. :'A Prediction-sampling-based Multilayer-structured Latent Factor Model for Accurate Representation to High-dimensional and Sparse Data'. IEEE Transactions on Neural Networks and Learning Systems, 2024, 35,(3),pp. 3845-3858.
[26]. Wu,D., He,Y., Luo, X. :'A Graph-incorporated Latent Factor Analysis Model for High-dimensional and Sparse Data'. IEEE Transactions on Emerging Topics in Computing, 2023, 11,(4),pp. 907-917.

[27]. Wu,D., Zhang,P., He,Y., et al. :'A Double-Space and Double-Norm Ensembled Latent Factor Model for Highly Accurate Web Service QoS Prediction', IEEE Transactions on Services Computing, 2023, 16,(2),pp. 802-814.

[28]. Wu,D., He,Q., Luo, X., et al. :'A Posterior-neighborhood-regularized Latent Factor Model for Highly Accurate Web Service QoS Prediction', IEEE Transactions on Services Computing, 2022, 15,(2),pp. 793-805.

[29]. Wu,D., He,Y., Luo, X., et al. :'A Latent Factor Analysis-based Approach to Online Sparse Streaming Feature Selection'. IEEE Transactions on Systems Man Cybernetics: Systems, 2022, 52(11),pp.6744-6758.

[30]. Wu,D., Shang,M., Luo, X., et al. :'An L1-and-L2-norm-oriented Latent Factor Model for Recommender Systems', IEEE Transactions on Neural Networks and Learning Systems, 2022, 33,(10),pp. 5775-5788.

[31]. Wu,D., Luo, X., He,Y., et al. :'A Prediction-sampling-based Multilayer-structured Latent Factor Model for Accurate Representation to High-dimensional and Sparse Data'. IEEE Transactions on Neural Networks and Learning Systems, 2024, 35,(3),pp.3845-3858.

[32]. Luo, X., Wang,Z., Shang,M. :'An Instance-frequency-weighted Regularization Scheme for Non-negative Latent Factor Analysis on High Dimensional and Sparse Data'. IEEE Transactions on Systems Man Cybernetics: Systems, 2021, 51,(6),pp.3522-3532.

[33]. Luo, X., Liu,Z., Shang,M., et al. :'Highly-Accurate Community Detection via Pointwise Mutual Information-Incorporated Symmetric Non-negative Matrix Factorization', IEEE Transactions on Network Science and Engineering, 2021, 8,(1), pp.463-476.

[34]. Li,Z., Li,S., Luo, X. :'An Overview of Calibration Technology of Industrial Robots', IEEE/CAA Journal of Automatica Sinica, 2021, 8,(1),pp.23-36.

[35]. Jin,L., Zhang,J., Luo, X., et al. :'Perturbed Manipulability Optimization in A Distributed Network of Redundant Robots'. IEEE Transactions on Industrial Electronics, 2021, 68, (8),pp. 7209-7220.

[36]. Luo, X., Zhou,M., Li,S.,et al. :'Non-negativity Constrained Missing Data Estimation for High-dimensional and Sparse Matrices from Industrial Applications'. IEEE Transactions on Cybernetics, 2020, 50,(5),pp. 1844-1855.

[37]. Li, Z. B., Li, S., Bamasag, O., Alhothali, A., and Luo, X.: 'Diversified regularization enhanced training for effective manipulator calibration.', IEEE Trans. Neural Netw. Learn. Syst., 2023, 34, (11), pp. 8778-8790

[38]. Wu, D., Bo Sun, and Shang, M. S.: 'Hyperparameter learning for deep learning-based recommender systems.', IEEE Trans. Serv. Comput., 2023, 16, (4), pp. 2699-2712

[39]. Yuan,Y., Luo, X., Shang,M.,et al. :'A Kalman-Filter-Incorporated Latent Factor Analysis Model for Temporally Dynamic Sparse Data'. IEEE Transactions on Cybernetics, 2023, 53,(9),pp. 5788-5801.

[40]. Luo, X., Liu,Z., Li,S., et al.:'A Fast Non-negative Latent Factor Model based on Generalized Momentum Method'. IEEE Transactions on Systems Man Cybernetics: Systems, 2021, 51,(1), pp. 610-620.

[41]. Li,W., Luo, X., Yuan,H. , et al.:'A Momentum-accelerated Hessian-vector-based Latent Factor Analysis Model'. IEEE Transactions on Services Computing, 2023, 16, (2), pp. 830-844.

[42]. Besta, M., Blach, N., Kubicek, A., et al.:'Graph of Thoughts: Solving Elaborate Problems with Large Language Models'. Proc. AAAI Conf. Artif. Intell., Vancouver, BC, Canada, February 2024, AAAI Press, 38,(16), pp. 17682–17690

[43]. Wei, J., Wang, X., Schuurmans, D., et al.:'Chain-of-Thought Prompting Elicits Reasoning in Large Language Models'. Proc. Adv. Neural Inf. Process. Syst., New Orleans, LA, USA, December 2022, Neural Inf. Process. Syst. Found., 35, pp. 24824–24837

[44]. Xie, Z., Jin, L., Luo, X. , et al.:'RNN for Repetitive Motion Generation of Redundant Robot Manipulators: An Orthogonal Projection Based Scheme'. IEEE Transactions on Neural Networks and Learning Systems, 2022, 33,(2), pp. 615-628.

[45]. Li, Z., Li, S., Luo, X.:'Using Quadratic Interpolated Beetle Antennae Search to Enhance Robot Arm Calibration Accuracy'. IEEE Robotics and Automation Letters, 2022, 7,(4), pp. 12046-12053.

[46]. Liu, Z., Yi, Y., Luo, X.:'A High-Order Proximity-Incorporated Nonnegative Matrix Factorization-based Community Detector'. IEEE Transactions on Emerging Topics in Computational Intelligence, 2023, 7,(3), pp. 700-714.

[47]. Tai, C. Y., Chen, Z., Zhang, T., et al.:'Exploring Chain of Thought Style Prompting for Text-to-SQL'. Proc. Conf. Empirical Methods in Natural Language Processing, Singapore, December 2023, Assoc. Comput. Linguistics, pp. 5376–5393

[48]. Wang, Z., Liu, Y., Luo, X. , et al.:'Large-Scale Affine Matrix Rank Minimization with a Novel Nonconvex Regularizer'. IEEE Transactions on Neural Networks and Learning Systems, 2022, 33,(9) , pp. 4661-4675.

[49]. Gan, Y., Chen, X., Purver, M.:'Exploring Underexplored Limitations of Cross-Domain Text-to-SQL Generalization'. Proc. Conf. Empirical Methods in Natural Language Processing, Punta Cana, Dominican Republic, November 2021, Assoc. Comput. Linguistics, pp. 8926–8931

[50]. Gao, T., Yao, X., Chen, D.:'SimCSE: Simple Contrastive Learning of Sentence Embeddings'. Proc. Conf. Empirical Methods in Natural Language Processing, Punta

Cana, Dominican Republic, November 2021, Assoc. Comput. Linguistics, pp. 6894–6910

[51]. Deng, S., Zhang, J. T., Wu, D., He, Y., Xie, X. P., and Wu, X. D.: 'A quantitative risk assessment model for distribution cyber physical system under cyber attack.',IEEE Trans. Industr. Inform., 2023, 19, (3), pp. 2899-2908

[52]. Yang, J., Wang, X. Q., Wang, G. Y., Zhang, Q. H., Zheng, N. G., Wu, D.: 'Fuzziness-based three-way decision with neighborhood rough sets under the framework of shadowed sets.', IEEE Trans. Fuzzy Syst., 2024, 32, (9), pp.4976-4988

[53]. You, D. L., Yan, H. G., Xiao, J. W., Chen, Z., Wu, D., Shen, L. M., and Wu, X. D.: 'Online learning for data streams with incomplete features and labels.', IEEE Trans. Knowl. Data Eng., 2024, 36, (9), pp.4820-4834

[54]. Chen, D. C., Li, S., Wu, Q., and Luo, X.: 'New disturbance rejection constraint for redundant robot manipulators: an optimization perspective.', IEEE Trans. Industr. Inform., 2020, 16, (4), pp. 2221-2232

[55]. Khan, A., Li, S., and Luo, X.: 'Obstacle avoidance and tracking control of redundant robotic manipulator: an RNN based metaheuristic approach.', IEEE Trans. Industr. Inform., 2020, 16, (7), pp. 4670-4680

[56]. Hu, L., Hu, P. W., Yuan, X. H., Luo, X., and You. Z. H.: 'Incorporating the coevolving information of substrates in predicting HIV-1 protease cleavage sites.', IEEE/ACM Transactions on Computational Biology and Bioinformatics, 2020, 17, (6), pp. 2017-2028

[57]. Achiam, J., Adler, S., Agarwal, S., et al.: 'GPT-4 Technical Report.', arXiv preprint arXiv:2303.08774, 2024

[58]. Bai, J., Bai, S., Chu, Y., et al.: 'Qwen Technical Report.', arXiv preprint arXiv:2309.16609, 2023

[59]. Brown, T., Mann, B., Ryder, N., et al.: 'Language Models are Few-Shot Learners.', Advances in Neural Information Processing Systems (NeurIPS), 2020, 33, pp. 1877 – 1901

[60]. Cao, R., Chen, L., Chen, Z., et al.: 'LGESQL: Line Graph Enhanced Text-to-SQL Model with Mixed Local and Non-Local Relations.', Proceedings of the 59th Annual Meeting of the Association for Computational Linguistics and the 11th International Joint Conference on Natural Language Processing (ACL-IJCNLP), 2021, pp. 2541 – 2555

[61]. DeepSeek-AI, Liu, A., Feng, B., et al.: 'DeepSeek-V3 Technical Report.', arXiv preprint arXiv:2412.19437, 2025

[62]. Demšar, J.: 'Statistical comparisons of classifiers over multiple data sets.', Journal of Machine Learning Research (JMLR), 2006, 7, pp. 1 – 30

[63]. Dong, X., Zhang, C., Ge, Y., et al.: 'C3: Zero-shot Text-to-SQL with ChatGPT.', arXiv preprint arXiv:2307.07306, 2023

[64]. .Dou, L., Gao, Y., Pan, M., et al.: 'UniSAr: A Unified Structure-Aware Autoregressive Language Model for Text-to-SQL.', arXiv preprint arXiv:2203.07781, 2022

[65]. Fu, H., Liu, C., Wu, B., et al.: 'CatSQL: Towards Real World Natural Language to SQL Applications.', Proceedings of the VLDB Endowment, 2023, 16, (6), pp. 1534 – 1547

[66]. Gao, Y., Liu, Y., Li, X., et al.: 'A Preview of XiYan-SQL: A Multi-Generator Ensemble Framework for Text-to-SQL.', arXiv preprint arXiv:2411.08599, 2025

[67]. Team GLM, Zeng, A., Xu, B., et al.: 'ChatGLM: A family of large language models from GLM-130B to GLM-4 all tools.', arXiv preprint arXiv:2406.12793, 2024

[68]. Gorti, S. K., Gofman, I., Liu, Z., et al.: 'MSc-SQL: Multi-Sample Critiquing Small Language Models for Text-to-SQL Translation.', Proceedings of the 2025 Conference of the North American Chapter of the Association for Computational Linguistics (NAACL), Long Papers, Vol. 1, 2025, pp. 2145 – 2160

[69]. Grattafiori, A., Dubey, A., Jauhri, A., et al.: 'The Llama 3 Herd of Models.', arXiv preprint arXiv:2407.21783, 2024

[70]. Haarnoja, T., Zhou, A., Abbeel, P., and Levine, S.: 'Soft Actor-Critic: Off-Policy Maximum Entropy Deep Reinforcement Learning with a Stochastic Actor.', Proceedings of the 35th International Conference on Machine Learning (ICML), 2018, Vol. 80, pp. 1856 – 1865